\documentclass[11pt]{article}

\usepackage[margin=1in]{geometry}
\usepackage[T1]{fontenc}
\usepackage[utf8]{inputenc}
\usepackage{lmodern}
\usepackage{amsmath,amssymb,amsfonts}
\usepackage{graphicx}
\usepackage{textcomp}
\usepackage{xcolor}
\usepackage{booktabs}
\usepackage{array}
\usepackage{tikz}
\usetikzlibrary{arrows.meta,positioning,calc}
\usepackage{url}
\usepackage{fancyvrb}
\usepackage[numbers,sort&compress]{natbib}
\usepackage{hyperref}
\hypersetup{hidelinks,
  pdftitle={Graph-of-Thoughts for LLM Causal Discovery with a Fairness-Pathway Case Study},
  pdfauthor={Nitish Nagesh, Elahe Khatibi, Thomas Hughes, Mahdi Bagheri, Pratik Gajane, Amir M. Rahmani},
  pdfkeywords={causal discovery, large language models, graph of thoughts, path-specific fairness, structure learning, clinical graphs}}
\DeclareFontFamily{OML}{lmtt}{\skewchar\font127}
\DeclareFontShape{OML}{lmtt}{m}{n}{<->ssub*cmm/m/it}{}
\DeclareFontShape{OML}{lmtt}{m}{it}{<->ssub*cmm/m/it}{}
\DeclareFontShape{OML}{lmtt}{bx}{n}{<->ssub*cmm/b/it}{}
\DeclareFontFamily{OMS}{lmtt}{\skewchar\font48}
\DeclareFontShape{OMS}{lmtt}{m}{n}{<->ssub*cmsy/m/n}{}
\DeclareFontShape{OMS}{lmtt}{bx}{n}{<->ssub*cmsy/b/n}{}
\DeclareFontFamily{OMX}{lmtt}{}
\DeclareFontShape{OMX}{lmtt}{m}{n}{<->ssub*cmex/m/n}{}
\newcounter{algocd}

\title{Graph-of-Thoughts for LLM Causal Discovery\\ with a Fairness-Pathway Case Study%
\thanks{Accepted as a workshop paper at the IEEE International Conference on Bioinformatics and Biomedicine (BIBM)~2026.}}

\author{
Nitish Nagesh$^{1}$\thanks{Corresponding author: \texttt{nnagesh1@uci.edu}.},
Elahe Khatibi$^{1}$,
Thomas Hughes$^{1}$,\\
Mahdi Bagheri$^{1}$,
Pratik Gajane$^{2}$,
Amir~M.~Rahmani$^{1}$
}

\date{}

\begin{document}
\maketitle

\begin{center}
\small
$^{1}$University of California, Irvine, Irvine, USA\\
$^{2}$University of Orl\'eans, Orl\'eans, France
\end{center}

\begin{abstract}
Methods that use large language models (LLMs) for causal discovery reason
locally, deciding one variable pair or one node at a time, so they never
compare alternative whole graphs.
We introduce GoT-CD, a Graph-of-Thoughts method that asks an LLM for several
complete candidate graphs from variable metadata alone, ranks them with a
deterministic validity score, merges them under a hard union constraint,
refines the merge, and projects the result to a DAG.
We evaluate on Asia, Child, and an Alzheimer's disease graph with 100
samples and GPT-4o-mini.
Over five seeds on Alzheimer's and Child, GoT-CD gives the highest mean F1
score and the lowest mean structural Hamming distance of the LLM methods, with 95\%
intervals of the difference excluding zero in seven of eight comparisons, and
it returns a DAG in every run while using three LLM requests per graph.
Replaying the logged runs without individual components attributes most of
the gain to generating and merging several complete candidates.
A case study on the effect of Sex on MOCA Score in the Alzheimer's graph
shows the limits of structural scores: the fairness verdict drawn from the
discovered graph disagrees with the reference verdict in twelve of fifteen LLM
runs, including all five GoT-CD runs, which mostly drop the edge out of Sex.
Graphs meant for a specific downstream analysis should therefore also be
checked on the pathways that analysis depends on.
\end{abstract}


\section{Introduction}
\label{sec:introduction}

% =============================================================================
% fig:got-pipeline — How GoT-CD differs from local LLM discovery (IEEE figure*).
% Requires: tikz; arrows.meta, positioning, calc (loaded in main.tex preamble)
% =============================================================================
\begin{figure}[!t]
\centering
\resizebox{\textwidth}{!}{%
\begin{tikzpicture}[
    every node/.style={align=center, font=\footnotesize},
    >={Stealth[length=2mm,width=1.6mm]},
    llm/.style={rectangle, draw=blue!55!black, fill=blue!6, rounded corners=2pt,
        inner sep=3pt, minimum height=7mm, text width=46mm},
    llmS/.style={rectangle, draw=blue!55!black, fill=blue!6, rounded corners=2pt,
        inner sep=3pt, minimum height=7mm, text width=20mm},
    thought/.style={rectangle, draw=blue!55!black, fill=white, rounded corners=1pt,
        inner sep=2pt, minimum height=6mm, minimum width=11mm},
    det/.style={rectangle, draw=black!55, fill=black!5, rounded corners=2pt,
        inner sep=3pt, minimum height=7mm, text width=46mm},
    post/.style={rectangle, draw=orange!60!black, fill=orange!10, rounded corners=2pt,
        inner sep=3pt, minimum height=9mm, text width=46mm},
    outW/.style={rectangle, draw=black!55, fill=white, dashed, rounded corners=2pt,
        inner sep=3pt, minimum height=9mm, text width=46mm},
    arr/.style={semithick, ->, draw=black!55},
    title/.style={font=\small\bfseries},
    note/.style={font=\scriptsize\itshape, text=black!60, text width=50mm}
]
% ---------------- (a) Pairwise ----------------
\begin{scope}[xshift=0cm]
\node[title] at (0,0.9) {(a) Pairwise querying};
\node[llm] (a1) at (0,0) {LLM: $v_1$ vs.\ $v_2$? \;($\to$, $\leftarrow$, or none)};
\node[llm] (a2) at (0,-1.1) {LLM: $v_1$ vs.\ $v_3$?};
\node (ad) at (0,-1.85) {$\vdots$};
\node[llm] (a3) at (0,-2.6) {LLM: $v_{n-1}$ vs.\ $v_n$?};
\node[outW] (ao) at (0,-4.1) {Output: all pairwise answers\\(no acyclicity check)};
\draw[arr] (a1) -- (a2); \draw[arr] (a2) -- (ad); \draw[arr] (ad) -- (a3);
\draw[arr] (a3) -- (ao);
\end{scope}

% ---------------- (b) LLM-BFS ----------------
\begin{scope}[xshift=5.8cm]
\node[title] at (0,0.9) {(b) LLM-BFS};
\node[llm] (b1) at (0,0) {LLM: which variables are roots?};
\node[llm] (b2) at (0,-1.1) {LLM: children of $u_1$ \;| edges so far};
\node (bd) at (0,-1.85) {$\vdots$};
\node[llm] (b3) at (0,-2.6) {LLM: children of $u_n$ \;| edges so far};
\node[outW] (bo) at (0,-4.1) {Output: union of expansions\\(no acyclicity check)};
\draw[arr] (b1) -- (b2); \draw[arr] (b2) -- (bd); \draw[arr] (bd) -- (b3);
\draw[arr] (b3) -- (bo);
\end{scope}

% ---------------- (c) GoT-CD ----------------
\begin{scope}[xshift=11.6cm]
\node[title] at (0,0.9) {(c) GoT-CD (ours)};
\node[llm] (c0) at (0,0) {\textsc{Generate}: $k$ complete graphs};
\node[thought] (g1) at (-1.5,-0.95) {$G_1$};
\node[thought] (g2) at (0,-0.95) {$G_2$};
\node[thought] (g3) at (1.5,-0.95) {$G_3$};
\draw[arr] (c0) -- (g1); \draw[arr] (c0) -- (g2); \draw[arr] (c0) -- (g3);
\node[det] (c1) at (0,-1.9) {\textsc{Score} (no LLM) + keep top $\min(2,k)$};
\draw[arr] (g1) -- (c1); \draw[arr] (g2) -- (c1); \draw[arr] (g3) -- (c1);
\node[llmS] (c2) at (-1.3,-2.9) {\textsc{Aggregate}\\[-1pt]\scriptsize $E\subseteq\mathcal{U}_{\mathrm{ret}}$};
\node[llmS] (c3) at (1.3,-2.9) {\textsc{Improve}\\[-1pt]\scriptsize edit pass};
\draw[arr] (c1) -- (c2); \draw[arr] (c2) -- (c3);
\node[post] (c4) at (0,-4.1) {Support filter + priority projection\\Output: DAG};
\draw[arr] (c3) -- (c4);
\end{scope}

% legend
\node[rectangle, draw=blue!55!black, fill=blue!6, minimum width=4mm, minimum height=3mm] (l1) at (3.2,-4.95) {};
\node[right=1mm of l1, font=\scriptsize] {LLM call};
\node[rectangle, draw=black!55, fill=black!5, minimum width=4mm, minimum height=3mm] (l2) at (6.2,-4.95) {};
\node[right=1mm of l2, font=\scriptsize] {deterministic step};
\node[rectangle, draw=orange!60!black, fill=orange!10, minimum width=4mm, minimum height=3mm] (l3) at (9.6,-4.95) {};
\node[right=1mm of l3, font=\scriptsize] {deterministic post-processing};
\end{tikzpicture}%
}
\caption{How GoT-CD differs from local LLM discovery.
(a)~Pairwise querying and (b)~LLM-BFS each make one local, final decision per
call.
(c)~GoT-CD proposes $k$ complete graphs, ranks them deterministically, merges
the best under a hard union constraint, edits the
merge, and projects the result to a DAG, preferring edges proposed by more
Generate candidates.
All three methods receive only variable names, descriptions, and a domain
preamble in their default form.}
\label{fig:got-pipeline}
\end{figure}

Causal discovery recovers a directed acyclic graph (DAG) of cause--effect
relations from observational
data~\cite{spirtes2000causation,pearl2009causality,glymour2019review}.
In clinical research such graphs underlie mechanism reasoning,
adjustment-set selection, and audits of how protected attributes influence
outcomes~\cite{prosperi2020causal,feuerriegel2024causal,brouillard_landscape_2025}.
Classical algorithms learn structure from the data alone through conditional
independence tests, score-based search, or continuous
optimization~\cite{spirtes2000causation,chickering2002optimal,zheng2018dags},
but they rest on parametric and faithfulness assumptions and lose power at
the small sample sizes typical of clinical studies.

Large language models (LLMs) offer a complementary source of evidence:
knowledge of what named variables mean and how they typically relate.
Pairwise querying asks the model whether each variable pair is causally
linked~\cite{kiciman2023causal} and assembles the answers into a full
graph~\cite{long2023causal}, while breadth-first traversal, which we call
LLM-BFS, expands one node at a time and passes previously discovered edges as
context~\cite{jiralerspong2024efficient}.
In parallel, prompting research has moved from single reasoning chains to
branching ones~\cite{wei2022chain,yao2023tree}, and
Graph-of-Thoughts~\cite{besta2024got} generates several candidate solutions,
scores them, and merges the best.

Existing LLM-based discovery nonetheless remains local.
Pairwise querying judges one pair at a time and sees only earlier decisions
involving that pair, and LLM-BFS commits to each node before later evidence
can revise the choice, so neither compares alternative whole graphs or checks
their global coherence.
It is therefore unclear whether reasoning over complete candidate graphs
improves prior-informed structure recovery.
A second question is whether structural accuracy is enough: a graph can score
well on edge-level $F_1$ yet omit, invent, or distort the specific pathway on
which a downstream analysis such as a path-specific fairness audit depends.

Here we introduce GoT-CD, which makes the complete edge set, rather than a
single edge or a node's children, the unit of Graph-of-Thoughts reasoning.
GoT-CD generates several full candidate graphs from variable metadata alone,
ranks them with a deterministic validity score, merges them so that the merge
cannot invent edges absent from every candidate, refines the result, and
projects it to a DAG.
Across five seeds, GoT-CD recovers structure better on average than local LLM
baselines on the Alzheimer's and Child graphs and returns a DAG in every run.
A case study on one fairness pathway in the Alzheimer's graph shows that this
advantage does not carry over to the pathway: GoT-CD most often drops every
edge out of Sex, whereas LLM-BFS and pairwise querying, with less accurate
graphs, keep the true path in some runs.

Our contributions are threefold.
First, we propose GoT-CD, a full-graph Graph-of-Thoughts proposer with
deterministic validity scoring, union-constrained aggregation, and
support-ordered DAG projection.
Second, a five-seed comparison with interval estimates indicates that GoT-CD
improves on local LLM baselines in $F_1$ and structural Hamming distance and
is DAG-valid in every run, and a replay of the logged runs without individual
components shows where the gain arises.
Third, a path-level case study shows that structural accuracy need not
predict whether a fairness-relevant pathway is recovered.
Code, prompts, data samples, all discovered graphs, and the LLM transcripts
are available at \url{https://anonymous.4open.science/r/GoT-CD-2026}.

%%%%%%%%%%%%%%%%%%%%%%%%%%%%%%%%%%%%%%%%%%%%%%%%%%%%%%%%%%%%%%%%%%%%%%%%%%%%%%%%

\section{Methods}
\label{sec:methods}

\subsection{Graph-of-Thoughts Causal Discovery}
\label{subsec:got-cd}

\textit{Problem and notation.}\ Let $V=\{v_1,\ldots,v_n\}$ be a finite set of
variables, each with a short natural-language description.
A domain preamble $D$ supplies dataset-specific clinical context.
A branch factor $k\geq 1$ sets how many candidate graphs \textsc{Generate}
produces.
A candidate graph, or thought, is a set of directed edges over $V\times V$.
An observational matrix $X\in\mathbb{R}^{m\times n}$ is used by the classical
baselines; GoT-CD uses it only to order the columns of the returned adjacency
matrix and, in a hybrid mode, to compute data-fit and correlation terms.
In the default mode, $X$ never reaches the LLM or the scorer.
The goal is to return a DAG adjacency matrix over $V$.

\textit{Design rationale.}\ Existing LLM discovery methods make one local
decision per call, as Fig.~\ref{fig:got-pipeline} illustrates: pairwise
querying decides one pair~\cite{kiciman2023causal,long2023causal}, LLM-BFS
fixes each node once it is expanded~\cite{jiralerspong2024efficient}, and
neither compares complete graphs or enforces acyclicity.
GoT-CD instead lets the LLM propose, merge, and edit whole graphs, while
deterministic code ranks candidates, filters edges, and projects the result
to a DAG.
Algorithm~\ref{alg:got-cd} lists the steps.

\textit{Information access.}\ The default LLM methods see only variable names,
descriptions, and the preamble; a hybrid GoT-CD mode admits the sample through
a Gaussian BIC term and correlation-weighted projection.

\begin{center}
\refstepcounter{algocd}\label{alg:got-cd}
\fbox{\begin{minipage}{0.95\textwidth}
\footnotesize
\textbf{Algorithm \thealgocd} GoT-CD causal discovery.\\[2pt]
\textbf{Input:} variables $V$ with descriptions, preamble $D$, branch factor
$k$, and an LLM.\\
\textbf{Output:} DAG adjacency matrix over $V$.
\begin{enumerate}\setlength{\itemsep}{1pt}\setlength{\parsep}{0pt}\setlength{\leftmargin}{1.2em}
\item $\{G_i\}_{i=1}^{k}\gets\textsc{Generate}(k)$ (one request, $k$ sampled completions)
\item Score each $G_i$; retain the top $\min(2,k)$
\item $\mathcal{U}_{\mathrm{ret}}\gets\bigcup$ retained candidates
\item $E\gets\textsc{Aggregate}$(retained), $E\subseteq\mathcal{U}_{\mathrm{ret}}$
\item $E^\star\gets\textsc{Improve}(E)$ (may add edges $\notin\mathcal{U}_{\mathrm{ret}}$)
\item Re-score; keep the single best candidate
\item $\mathcal{U}_{\mathrm{gen}}\gets\bigcup_{i=1}^{k}G_i$
\item $E\gets(E^\star\cap\mathcal{U}_{\mathrm{gen}})\cup\{e\in E^\star\setminus\mathcal{U}_{\mathrm{gen}}:\textrm{cycle-safe}\}$
\item If $E=\emptyset$, set $E$ to the union of the top $\min(2,k)$ Generate candidates
\item Project $E$ to a DAG in priority order (Generate frequency; hybrid adds $|\mathrm{corr}|$)
\item Return the adjacency matrix of $E$
\end{enumerate}
\end{minipage}}
\end{center}

\textit{Generate.}\ \textsc{Generate} sends one prompt and samples $k$
independent completions from it.
Each completion is parsed into one candidate graph $G_i$ over all of $V$.
The purpose of this step is diversity: several complete hypotheses that can
be ranked and merged, rather than a single irrevocable proposal.

\textit{Score.}\ \textsc{Score} ranks candidates with a deterministic
function and never calls the LLM.
The default validity heuristic adds $1$ per edge, adds $3$ if the graph is
acyclic and subtracts $4$ otherwise, and applies a density penalty to a
graph with $|E|$ edges: it subtracts $0.1(0.5n-|E|)$ when $|E|<0.5n$ and
$0.5(|E|-2.5n)$ when $|E|>2.5n$.
An empty graph receives a score of $-1$.
In hybrid mode, the score additionally includes $-0.01\cdot\mathrm{BIC}$,
where BIC comes from a Gaussian least-squares fit of each node on its
candidate parents.
The score is never an LLM self-rating, and \textsc{KeepBestN} passes the top
$\min(2,k)$ candidates to \textsc{Aggregate}.

\textit{Aggregate.}\ \textsc{Aggregate} asks the LLM to merge the retained
candidates, shown with their scores, into one edge set $E$.
Let $\mathcal{U}_{\mathrm{ret}}$ denote the union of the retained candidates.
The prompt instructs the LLM to output only edges from
$\mathcal{U}_{\mathrm{ret}}$, and the parser removes any edge outside it, so
$E\subseteq\mathcal{U}_{\mathrm{ret}}$ holds by construction.
This constraint prevents the merge step from introducing a plausible-sounding
edge that no retained candidate proposed.

\textit{Improve.}\ \textsc{Improve} sends $E$ back to the LLM for a
conservative edit pass that may delete edges or add missing direct edges,
including edges outside $\mathcal{U}_{\mathrm{ret}}$.
We denote its output by $E^\star$.
This step repairs merges that are too sparse or locally inconsistent, while
the post-processing below still controls any new edge.

\textit{Post-processing.}\ Let $\mathcal{U}_{\mathrm{gen}}=\bigcup_{i=1}^{k}G_i$
be the union of all $k$ Generate candidates.
GoT-CD keeps every edge of $E^\star$ that lies in $\mathcal{U}_{\mathrm{gen}}$.
It then admits each edge proposed only by \textsc{Improve}, one at a time,
provided that the edge does not close a directed cycle with the edges kept so
far.
If no edge survives, GoT-CD falls back to the union of the top $\min(2,k)$
Generate candidates by score.

\textit{Priority DAG projection.}\ The surviving edges are sorted by priority
and inserted greedily, and an edge is skipped if it would close a directed
cycle.
An edge's default priority is the number of Generate candidates that contain
it, and an edge proposed only by \textsc{Improve} receives a priority of
$0.25$.
In hybrid mode, the absolute Pearson correlation between the two variables
is added to that priority, so cycle conflicts favor edges that are both often
proposed and supported by the data.
Ties are broken by the lexicographic order of source and target names.
This projection guarantees a DAG, and the outcome depends on explicit support
rather than on the order in which the LLM listed the edges.

\textit{Prompting and parsing.}\ Every prompt contains the domain preamble
$D$, a variable block with one \texttt{name: description} line per variable,
and a stage-specific checklist; Fig.~\ref{fig:prompts} reproduces the
complete \textsc{Generate} prompt for Alzheimer's and the checklists of
\textsc{Aggregate} and \textsc{Improve}.
Each response must list its edges inside \texttt{<Answer>...</Answer>}, one
per line as \texttt{A -> B}.
The parser matches declared variable names case-insensitively and
longest-first, so that multi-word names such as Brain Volume are not
truncated, and it discards self-loops and undeclared names.

\textit{Hyperparameters.}\ All reported experiments use temperature $0.7$,
branch factor $k{=}3$, a limit of $4096$ tokens per response, which
suffices to emit a full edge list, and the \texttt{gpt-4o-mini} snapshot of
18 July 2024.
We additionally vary $k\in\{1,3,5\}$ on an independent grid.
A run issues three requests whatever the number of variables:
\textsc{Generate} (all $k$ completions in one request), \textsc{Aggregate},
and \textsc{Improve}.

\begin{figure}[!t]
\centering
\newsavebox{\gotcdpromptbox}
\begin{lrbox}{\gotcdpromptbox}
\begin{minipage}{516pt}
\setlength{\textwidth}{516pt}
\fvset{fontsize={\fontsize{7}{8}\selectfont},frame=single,framesep=3pt,rulecolor=\color{black!40}}
\begin{minipage}[t]{0.495\textwidth}
\textbf{\footnotesize(a) \textsc{Generate}, Alzheimer's, $k{=}3$ (complete prompt)}
\begin{Verbatim}
You are an expert on alzheimer's disease in patients. Apply your
knowledge to a challenging real-world dataset from the Alzheimer's
Disease Neuroimaging Initiative (ADNI) (Petersen et al., 2010). We
consider clinical and radiological phenotypes of AD and how
variables might impact a brain Magnetic Resonance Image (MRI) for a
given participant.
You are a helpful assistant to medical experts in diagnosis of
alzheimer's disease. Our goal is to construct a causal graph
between the following variables.

Variables (use these names exactly):
APOE4: Expression level of APOE4 gene
Sex: Biological sex of the patient
Age: Age of the patient
Education: Educational attainment (years)
AV45: Beta amyloid protein level measured by Florbetapir F18
P-tau: Phosphorylated-tau deposition
Brain Volume: Total brain matter volume of the patient
Ventricular Volume: Total ventricular volume of the patient
MOCA Score: Montreal Cognitive Assessment Score
Brain MRI: MRI image of the brain
Slice Number: Slice number of the MRI image

Task: Propose a directed causal graph over ALL of these variables.
Checklist before answering:
1. Direct causes only (not mere associations or reverse effects).
2. Acyclic: if A causes B, do not also assert a directed path
   back to A.
3. Sparse: include an edge only when a clear domain mechanism
   supports it.
4. Exact names: copy variable names from the list above, never
   paraphrase.
You will produce 3 distinct candidate graph(s).

Reason briefly about mechanisms, then output ONLY the edge list
inside <Answer>...</Answer>, one edge per line as:
A -> B
\end{Verbatim}
\end{minipage}\hfill
\begin{minipage}[t]{0.495\textwidth}
\textbf{\footnotesize(b) \textsc{Aggregate} (after preamble and variables)}
\begin{Verbatim}
Candidate graphs to aggregate:
Option 0 (score=<s0>):
<edge list>
Option 1 (score=<s1>):
<edge list>

Merge complementary edges into one coherent causal DAG.
Hard rules:
1. Output ONLY edges that already appear in at least one Option
   above.
2. Never invent new edges, even if they seem plausible.
3. If two options conflict, keep the higher-scoring option's edge
   and drop the other.
4. Prefer an acyclic, sparse union over a dense compromise.
5. Use only: <variable names>

After checking the rules, output the merged edge list in
<Answer>...</Answer> as:
A -> B
\end{Verbatim}
\textbf{\footnotesize(c) \textsc{Improve} (after preamble and variables)}
\begin{Verbatim}
Current candidate causal graph:
<edge list>

Improve this graph with a conservative edit pass:
1. Remove edges that are correlational, reverse-causal, or weakly
   justified.
2. Add a missing DIRECT edge only when the mechanism is clear
   among listed variables.
3. Enforce a DAG (no directed cycles).
4. Prefer sparse domain-plausible structure over completeness.
5. Use only: <variable names>

Self-check: every added edge must be a direct cause, and the final
list must be acyclic.
Then output the improved edge list in <Answer>...</Answer> as:
A -> B
\end{Verbatim}
\end{minipage}

\vspace{2pt}
\begin{minipage}[t]{0.495\textwidth}
\textbf{\footnotesize(d) Pairwise, one request per variable pair $\{X,Y\}$}
\begin{Verbatim}
[system] You are an expert on Alzheimer's disease.
[user] You are a helpful assistant to an expert on Alzheimer's
disease. Here is a description of the causal variables in this
causal graph: <variables>
Here are the causal relationships you know so far: <edges>
We are interested in the causal relationship between "X" and "Y".
Which cause-and-effect relationship is more likely?
A. "X" causes "Y". B. "Y" causes "X".
C. There is no causal relationship between "X" and "Y".
Let's work this out in a step by step way to be sure that we have
the right answer. Then provide your final answer within the tags
<Answer>A/B/C</Answer>.
\end{Verbatim}
\end{minipage}\hfill
\begin{minipage}[t]{0.495\textwidth}
\textbf{\footnotesize(e) LLM-BFS, first request, then one per expanded node $X$}
\begin{Verbatim}
<preamble and variables as in (a)>
Now you are going to use the data to construct a causal graph.
You will start with identifying the variable(s) that are
unaffected by any other variables.
...
Given <roots> is(are) not affected by any other variable and the
following causal relationships.
<A causes B, C>
Select variables that are caused by X.
Think step by step. Then, provide your final answer (variable
names only) within the tags <Answer>...</Answer>, seperated
by ", ".
\end{Verbatim}
\end{minipage}
\end{minipage}
\end{lrbox}
\resizebox{\textwidth}{!}{\usebox{\gotcdpromptbox}}
\caption{Prompts used in all reported runs, reproduced from the released code;
line breaks were added for layout, \texttt{<...>} marks run-time
insertions, and spelling errors in the original prompts are kept. GoT-CD and LLM-BFS share the dataset preamble shown in (a); the
pairwise baseline uses the shorter system and user prompts in (d). No prompt
contains observational data.}
\label{fig:prompts}
\end{figure}

\subsection{Datasets and reference graphs}
\label{subsec:datasets}

We evaluate on three reference graphs, drawn in Fig.~\ref{fig:reference-graphs}.
Asia, with $8$ nodes and $8$ edges, is the network of
\cite{lauritzen1988local} relating a visit to Asia and smoking to
tuberculosis, lung cancer, bronchitis, and dyspnea.
Child, with $20$ nodes and $25$ edges, is the network of
\cite{spiegelhalter1992learning} for diagnosing congenital heart disease in
newborns.
For both, we take the reference DAG and conditional probability tables from
the published networks and draw $100$ observations under a fixed seed, so
that every method sees the same data.
Both networks are widely reproduced and may appear in LLM pretraining
corpora~\cite{srivastava2025realizing,brouillard_landscape_2025}.
The Alzheimer's disease graph, with $11$ nodes and $19$ edges, comes
from~\cite{abdulaal2024causal} and was later curated as a less exposed
benchmark together with linear-Gaussian observational data generated from
it~\cite{srivastava2025realizing}; we use that graph as the reference and a
fixed-seed subsample of $100$ rows of the released data; its variables are
listed in Fig.~\ref{fig:prompts}a.
LLM methods never see the observational samples in their prompts.

% =============================================================================
% fig:reference-graphs — ground-truth DAGs (Asia, Alzheimer's, Child).
% Edges transcribed from the published reference graphs; node names verbatim.
% Requires: tikz; arrows.meta (loaded in main.tex preamble)
% =============================================================================
\begin{figure}[!t]
\centering
\resizebox{0.86\textwidth}{!}{%
\begin{tikzpicture}[yscale=0.72,
    >={Stealth[length=1.8mm,width=1.4mm]},
    v/.style={rectangle, rounded corners=2pt, draw=black!60, fill=black!4,
        inner sep=2pt, minimum height=5mm, font=\scriptsize},
    vp/.style={v, draw=red!70!black, fill=red!6, line width=0.8pt},
    e/.style={->, draw=black!55, semithick},
    ep/.style={->, draw=red!75!black, line width=1.1pt},
    title/.style={font=\small\bfseries}
]
% ---------------- (a) Asia: 8 nodes, 8 edges ----------------
\begin{scope}[xshift=0cm]
\node[title] at (1.3,0.8) {(a) Asia (8 nodes, 8 edges)};
\node[v]  (asia)   at (0,0)      {asia};
\node[v]  (smoke)  at (2.6,0)    {smoke};
\node[v]  (tub)    at (0,-1.3)   {tub};
\node[v]  (lung)   at (1.8,-1.3) {lung};
\node[v]  (bronc)  at (3.4,-1.3) {bronc};
\node[v]  (either) at (0.9,-2.6) {either};
\node[v]  (xray)   at (0,-3.9)   {xray};
\node[v]  (dysp)   at (2.4,-3.9) {dysp};
\draw[e]  (asia) -- (tub);
\draw[e]  (tub) -- (either);
\draw[e]  (smoke) -- (lung);
\draw[e]  (smoke) -- (bronc);
\draw[e]  (lung) -- (either);
\draw[e]  (bronc) -- (dysp);
\draw[e]  (either) -- (xray);
\draw[e]  (either) -- (dysp);
\end{scope}

% ---------------- (b) Alzheimer's: 11 nodes, 19 edges ----------------
\begin{scope}[xshift=6.2cm]
\node[title] at (3.4,0.8) {(b) Alzheimer's (11 nodes, 19 edges)};
\node[v]  (apoe)  at (0.4,0)    {APOE4};
\node[v]  (age)   at (3.0,0)    {Age};
\node[vp] (sex)   at (5.2,0)    {Sex};
\node[v]  (edu)   at (6.9,0)    {Education};
\node[v]  (av45)  at (0.8,-1.2) {AV45};
\node[v]  (ptau)  at (1.6,-2.3) {P-tau};
\node[vp] (bv)    at (3.9,-3.2) {Brain Volume};
\node[v]  (vv)    at (1.4,-4.3) {Ventricular Volume};
\node[vp] (moca)  at (5.9,-4.3) {MOCA Score};
\node[v]  (mri)   at (3.4,-5.4) {Brain MRI};
\node[v]  (slice) at (6.0,-5.4) {Slice Number};
\draw[e]  (apoe) -- (av45);
\draw[ep] (sex) -- (bv);
\draw[e]  (sex) -- (vv);
\draw[e]  (age) -- (av45);
\draw[e]  (age) -- (ptau);
\draw[e]  (age) -- (bv);
\draw[e]  (age) -- (moca);
\draw[e]  (age) -- (vv);
\draw[e]  (edu) -- (moca);
\draw[e]  (slice) -- (mri);
\draw[e]  (av45) -- (ptau);
\draw[e]  (av45) -- (bv);
\draw[e]  (ptau) -- (bv);
\draw[e]  (ptau) to[out=-60,in=190] (moca);
\draw[e]  (ptau) -- (vv);
\draw[ep] (bv) -- (moca);
\draw[e]  (bv) -- (vv);
\draw[e]  (bv) -- (mri);
\draw[e]  (vv) -- (mri);
\end{scope}

% ---------------- (c) Child: 20 nodes, 25 edges ----------------
\begin{scope}[yshift=-7.5cm, xshift=1.2cm]
\node[title] at (6.2,0.8) {(c) Child (20 nodes, 25 edges)};
\node[v]  (ba)    at (6.0,0)     {BirthAsphyxia};
\node[v]  (dis)   at (6.0,-1.1)  {Disease};
\node[v]  (duct)  at (0.8,-2.3)  {DuctFlow};
\node[v]  (cm)    at (3.0,-2.3)  {CardiacMixing};
\node[v]  (lp)    at (5.2,-2.3)  {LungParench};
\node[v]  (lf)    at (7.4,-2.3)  {LungFlow};
\node[v]  (sick)  at (9.6,-2.3)  {Sick};
\node[v]  (lvh)   at (11.8,-2.3) {LVH};
\node[v]  (hd)    at (0.8,-3.5)  {HypDistrib};
\node[v]  (hyp)   at (3.0,-3.5)  {HypoxiaInO2};
\node[v]  (co2)   at (4.8,-3.5)  {CO2};
\node[v]  (cx)    at (6.6,-3.5)  {ChestXray};
\node[v]  (gr)    at (8.4,-3.5)  {Grunting};
\node[v]  (agec)  at (10.3,-3.5) {Age};
\node[v]  (lvhr)  at (11.8,-3.5) {LVHreport};
\node[v]  (lbo)   at (1.6,-4.7)  {LowerBodyO2};
\node[v]  (ruq)   at (3.5,-4.7)  {RUQO2};
\node[v]  (co2r)  at (5.1,-4.7)  {CO2Report};
\node[v]  (xr)    at (6.8,-4.7)  {XrayReport};
\node[v]  (grr)   at (8.6,-4.7)  {GruntingReport};
\draw[e]  (ba) -- (dis);
\draw[e]  (dis) -- (duct);
\draw[e]  (dis) -- (cm);
\draw[e]  (dis) -- (lp);
\draw[e]  (dis) -- (lf);
\draw[e]  (dis) -- (sick);
\draw[e]  (dis) -- (lvh);
\draw[e]  (dis) -- (agec);
\draw[e]  (duct) -- (hd);
\draw[e]  (cm) -- (hd);
\draw[e]  (cm) -- (hyp);
\draw[e]  (lp) -- (hyp);
\draw[e]  (lp) -- (co2);
\draw[e]  (lp) -- (cx);
\draw[e]  (lp) -- (gr);
\draw[e]  (lf) -- (cx);
\draw[e]  (sick) -- (gr);
\draw[e]  (sick) -- (agec);
\draw[e]  (lvh) -- (lvhr);
\draw[e]  (hd) -- (lbo);
\draw[e]  (hyp) -- (lbo);
\draw[e]  (hyp) -- (ruq);
\draw[e]  (co2) -- (co2r);
\draw[e]  (cx) -- (xr);
\draw[e]  (gr) -- (grr);
\end{scope}
\end{tikzpicture}%
}
\caption{Reference DAGs drawn from the ground-truth adjacency matrices with
the original variable names: (a)~Asia~\cite{lauritzen1988local},
(b)~Alzheimer's disease~\cite{abdulaal2024causal}, and
(c)~Child~\cite{spiegelhalter1992learning}.
Red marks the Sex$\to$Brain Volume$\to$MOCA Score pathway used in the
fairness-pathway case study.}
\label{fig:reference-graphs}
\end{figure}

\subsection{Baseline methods}
\label{subsec:baselines}

Four classical baselines represent data-driven discovery:
PC~\cite{spirtes2000causation} and GES~\cite{chickering2002optimal} as
constraint-based and score-based methods, and NOTEARS~\cite{zheng2018dags}
and DAGMA-linear~\cite{bello2022dagma} as continuous-optimization methods.
PC uses Fisher-$Z$ tests at significance level $0.05$, GES the local BIC
score, NOTEARS an $\ell_2$ loss with $\lambda_1{=}0.01$ and weight threshold
$0.3$, and DAGMA-linear $\lambda_1{=}0.01$.
PC and GES return a completed partially directed graph (CPDAG), and we
encode each undirected edge as a pair of opposite directed edges, so an
unresolved orientation makes the output non-DAG.
With $100$ samples the GES and NOTEARS outputs can change with numerical-library
versions, so the released code includes every reported graph.
Two LLM baselines use the same backbone and temperature and likewise receive
no observational samples: pairwise querying~\cite{kiciman2023causal}, with
one request per variable pair, and LLM-BFS~\cite{jiralerspong2024efficient},
with one request to find the root variables and one per expanded node.
LLM-BFS shares the GoT-CD domain preamble, and pairwise querying uses the
shorter one-sentence expert prompt (Fig.~\ref{fig:prompts}).
We ran variants of both that also receive correlation statistics in
preliminary experiments but omit them, so that the LLM comparison isolates
reasoning topology rather than access to summary statistics.

\subsection{Structural evaluation}
\label{subsec:structural-eval}

We compare recovered graphs with the reference by directed-edge precision,
recall, and $F_1$, following~\cite{jiralerspong2024efficient}; an undirected
CPDAG edge therefore counts as two predicted edges, at most one of them
correct.
We report DAG validity separately, because a cyclic or partially directed
graph is not a valid causal DAG for many downstream uses, and resolving it is
a modeling choice absent from the method's own output.
For every DAG-valid estimate we also report the structural Hamming distance
(SHD), in which a reversed edge counts once and each missing or extra edge
counts once; SHD is omitted for non-DAG outputs.
Both metrics are exact against the supplied reference DAG and do not credit
recovery of its Markov equivalence class.

\subsection{Fairness-pathway case study}
\label{subsec:fairness-methods}

Path-specific fairness asks which routes carry the effect of a protected
attribute on an outcome~\cite{chiappa2019path,wu2019pcfairness}.
We check whether each discovered Alzheimer's graph keeps the route by which
Sex affects MOCA Score, which in the reference graph is the single path
Sex$\to$Brain Volume$\to$MOCA Score.
Given a DAG and the $100$-row sample used in discovery, we fit a linear
structural equation model by regressing each node on its parents; the effect
of a path is the product of coefficients along it, and the total effect is the
sum over paths~\cite{pearl2001direct,zhou2023tracing}.
The only non-DAG output on Alzheimer's, from PC, is made acyclic by
repeatedly deleting the last edge of a detected cycle.
We resample the rows $1000$ times and take the 95\% percentile interval of the
total effect; the verdict is negative or positive when that interval excludes
zero and none otherwise, including graphs with no path, and a flip is any
disagreement with the reference verdict.
The case study is exploratory and is not a clinical validation.

%%%%%%%%%%%%%%%%%%%%%%%%%%%%%%%%%%%%%%%%%%%%%%%%%%%%%%%%%%%%%%%%%%%%%%%%%%%%%%%%

\section{Results}
\label{sec:results-n100}

Because classical and LLM methods have different information access, we
report them as separate groups and hybrid GoT-CD separately from both.

\begin{table}[!t]
\centering
\caption{Structural recovery with $100$ samples and seed $0$.
Each cell gives $F_1$ and, for DAG-valid outputs, SHD in parentheses; column
headers give true edge counts, and $^\dagger$ marks a non-DAG output (for PC
and GES, a CPDAG with undirected edges).
Bold marks the best DAG-valid $F_1$ within each family.}
\label{tab:cross-dataset-f1}
\footnotesize
\setlength{\tabcolsep}{3pt}
\begin{tabular}{@{}lccc c@{}}
\toprule
\textbf{Method} & \textbf{Asia} & \textbf{Child} & \textbf{Alzheimer's} &
\textbf{DAG-valid} \\
 & $8$ edges & $25$ edges & $19$ edges & \\
\midrule
\multicolumn{5}{l}{\textit{Classical, data-driven}} \\
PC           & $0.333^\dagger$ & $0.383^\dagger$ & $0.312^\dagger$ & 0/3 \\
GES          & $0.235^\dagger$ & $0.302^\dagger$ & $\mathbf{0.650}$~(12) & 1/3 \\
NOTEARS      & $0.308$~(8) & $0.226$~(33) & $0.426$~(21) & 3/3 \\
DAGMA-linear & $\mathbf{0.333}$~(7) & $\mathbf{0.275}$~(30) & $0.522$~(18) & 3/3 \\
\midrule
\multicolumn{5}{l}{\textit{LLM, metadata only}} \\
LLM-pairwise & $0.519$~(13) & $\mathbf{0.226}$~(118) & $0.488$~(21) & 3/3 \\
LLM-BFS      & $0.600$~(8) & $0.160$~(40) & $0.649$~(13) & 3/3 \\
GoT-CD       & $\mathbf{0.750}$~(4) & $0.222$~(32) & $\mathbf{0.757}$~(8) & 3/3 \\
\bottomrule
\end{tabular}
\end{table}

\subsection{Structural recovery in the seed-0 run}
\label{subsec:cross-f1}

In the seed-$0$ run of every method, GoT-CD gives the best LLM $F_1$ on Asia
and Alzheimer's and the lowest SHD of any DAG-valid output on both; on
Alzheimer's it recovers $14$ of the $19$ true edges while predicting $18$.
Among DAG-valid classical outputs, GES is strongest on Alzheimer's and
DAGMA-linear on Asia and Child; on Child, DAGMA-linear exceeds every LLM
method.
Child is hard for every LLM method: pairwise querying predicts $134$ edges
against $25$ true ones, and GoT-CD predicts $20$.

\subsection{Five-seed structural recovery}
\label{subsec:multiseed}

Because sampling at temperature $0.7$ makes every LLM run stochastic, we
repeat the three LLM methods over five seeds on Alzheimer's and Child;
classical methods are deterministic given the sample.
A seed fixes the local random state, such as the order in which pairwise
querying presents each pair, but the API sampling itself is not seeded, so
each seed is an independent repeat.
GoT-CD has the highest mean $F_1$ and the lowest mean SHD of the three on
both datasets and is the only method that is DAG-valid in all ten runs, while
pairwise querying has the smallest $F_1$ spread across seeds.
Because the repeats are independent, we compare GoT-CD with each baseline as
two samples, using the mean difference, a percentile-bootstrap 95\% interval,
and an exact permutation test.
The interval excludes zero in seven of eight comparisons.
GoT-CD raises $F_1$ over pairwise querying by $0.105$ $[0.029, 0.188]$ on
Alzheimer's and $0.075$ $[0.039, 0.107]$ on Child, raises it over LLM-BFS by
$0.170$ $[0.118, 0.218]$ on Child, and lowers SHD against both baselines on
both datasets, for example by $3.6$ $[1.2, 6.2]$ against LLM-BFS on
Alzheimer's.
The exception is $F_1$ against LLM-BFS on Alzheimer's, $0.038$
$[-0.060, 0.145]$, where the two methods are not separable.
The permutation $p$-values of the other seven lie between $0.008$ and
$0.048$; with five runs per method none survives a Holm correction over all
eight, so we read them as consistent but not confirmatory evidence.
In one LLM-BFS run on Alzheimer's the model declared every variable
exogenous and returned an empty graph; we repeated that run and report the
repeat.
Compared with seed means, the seed-$0$ run overstates GoT-CD on Alzheimer's
and on Asia, where three repeated runs average $0.658$, and understates it on
Child, so we read the seed-$0$ table as a single draw rather than a
population estimate.

\begin{table}[t]
\centering
\caption{Five-seed structural recovery of the LLM methods.
Entries are mean${\pm}$sd over seeds $0$--$4$; SHD averages DAG-valid runs.
The last column counts DAG-valid runs out of five.}
\label{tab:multiseed}
\footnotesize
\setlength{\tabcolsep}{3pt}
\begin{tabular}{@{}lccc@{}}
\toprule
\textbf{Method} & $F_1$ & SHD & \textbf{DAG-valid} \\
\midrule
\multicolumn{4}{l}{\textit{Alzheimer's ($19$ edges)}} \\
LLM-pairwise & $0.586{\pm}0.047$ & $17.8{\pm}1.1$ & 5/5 \\
LLM-BFS      & $0.653{\pm}0.096$ & $13.0{\pm}2.3$ & 5/5 \\
GoT-CD       & $\mathbf{0.691{\pm}0.091}$ & $\mathbf{9.4{\pm}2.1}$ & 5/5 \\
\midrule
\multicolumn{4}{l}{\textit{Child ($25$ edges)}} \\
LLM-pairwise & $0.223{\pm}0.027$ & $111.7{\pm}15.0$ & 3/5 \\
LLM-BFS      & $0.127{\pm}0.054$ & $37.5{\pm}9.5$ & 4/5 \\
GoT-CD       & $\mathbf{0.298{\pm}0.035}$ & $\mathbf{27.6{\pm}2.5}$ & 5/5 \\
\bottomrule
\end{tabular}
\end{table}

\subsection{Component analysis}
\label{subsec:ablation}

Each GoT-CD transcript records the $k$ \textsc{Generate} completions and the
\textsc{Aggregate} and \textsc{Improve} answers, so a run can be replayed
without a component on exactly the same LLM samples; replaying the full
pipeline reproduces all ten stored graphs.
The gain comes mainly from generating and merging several complete
candidates.
A single \textsc{Generate} candidate, projected to a DAG, averages $0.654$ on
Alzheimer's and $0.215$ on Child, and full GoT-CD is better in eight of the ten
paired runs (exact sign-flip $p{=}0.020$).
Dropping \textsc{Improve}, or replacing \textsc{Aggregate} and
\textsc{Improve} with a majority vote over the candidates, keeps part of the
gain, and full GoT-CD is better in six of ten runs against each.
The \textsc{Improve} answers were already acyclic in all ten runs, so the
priority projection changed no graph here; it guarantees a DAG rather than
adding accuracy.

\begin{table}[t]
\centering
\caption{Component analysis by replaying the ten logged GoT-CD runs.
Entries are mean $F_1$ over five seeds; a single candidate is averaged over
the three \textsc{Generate} completions of each run.
B/T/W counts the ten paired runs in which full GoT-CD is better, tied, or
worse, and $p$ is the exact two-sided sign-flip $p$-value.}
\label{tab:ablation}
\footnotesize
\setlength{\tabcolsep}{3pt}
\begin{tabular}{@{}lcccc@{}}
\toprule
\textbf{Variant} & \textbf{Alz.} & \textbf{Child} & \textbf{B/T/W} & $p$ \\
\midrule
Full GoT-CD                      & $\mathbf{0.691}$ & $\mathbf{0.298}$ & -- & -- \\
Without \textsc{Improve}          & $0.680$ & $0.256$ & 6/2/2 & $0.109$ \\
Majority vote of candidates       & $0.675$ & $0.235$ & 6/1/3 & $0.156$ \\
Single \textsc{Generate} candidate & $0.654$ & $0.215$ & 8/0/2 & $0.020$ \\
\bottomrule
\end{tabular}
\end{table}

\subsection{Fairness-pathway case study on Alzheimer's}
\label{sec:fairness}

\begin{table}[t]
\centering
\caption{Sex$\to$MOCA Score pathway on Alzheimer's.
LLM entries count runs out of five; classical methods are deterministic, so
their entries count one run.
True path: the graph contains Sex$\to$Brain Volume$\to$MOCA Score.
No path: no path from Sex to MOCA Score.
Direct: the graph contains Sex$\to$MOCA Score, which the reference lacks.
None and Reversed: the bootstrap verdict differs from the reference by
finding no effect or an effect of the opposite sign.
TE: range of the total effect over runs; the reference is $-0.572$.}
\label{tab:alz-pathway}
\footnotesize
\setlength{\tabcolsep}{3pt}
\begin{tabular}{@{}lcccccc@{}}
\toprule
& \textbf{True} & \textbf{No} & & & & \\
\textbf{Method} & \textbf{path} & \textbf{path} & \textbf{Direct} & \textbf{None} & \textbf{Reversed} & \textbf{TE} \\
\midrule
LLM-pairwise & 1 & 0 & 4 & 3 & 1 & $-0.604$ to $0.520$ \\
LLM-BFS      & 2 & 3 & 2 & 3 & 0 & $-0.782$ to $0$ \\
GoT-CD       & 0 & 4 & 1 & 5 & 0 & $0$ to $0.053$ \\
\midrule
GES          & 1 & 0 & 0 & 0 & 0 & $-0.843$ \\
PC           & 0 & 1 & 0 & 1 & 0 & $0$ \\
NOTEARS      & 0 & 1 & 0 & 1 & 0 & $0$ \\
DAGMA-linear & 0 & 1 & 0 & 1 & 0 & $0$ \\
\bottomrule
\end{tabular}
\end{table}

In the reference graph the effect of Sex on MOCA Score is $-0.572$, with a
95\% interval from $-0.695$ to $-0.440$, so the reference verdict is negative.
Only three of the fifteen LLM runs reproduce it, and each contains the true
path.
GoT-CD keeps Brain Volume$\to$MOCA Score in every run but gives Sex no edges
in four, so the estimated effect is exactly zero, which reads as a
fair structure unless path recall is checked; in the fifth run a direct edge
carries an effect whose interval includes zero.
Pairwise querying instead adds a direct Sex$\to$MOCA edge in four runs, and
LLM-BFS drops the pathway in three runs and in the other two keeps it but
adds a direct edge that offsets part of the effect.
Among classical methods only GES keeps the true path, but it finds nine
routes from Sex to MOCA Score whose absolute effects sum to about seven times
the reference effect; the true path itself has the wrong sign, and the
correct negative verdict arises from cancellation among the routes.

\subsection{Branch factor and hybrid scoring}
\label{subsec:robustness}

On a separate seed-$0$ grid, GoT-CD is DAG-valid for every branch factor
$k\in\{1,3,5\}$.
Raising $k$ from $1$ to $5$ improves $F_1$ on Asia from $0.556$ to $0.667$
and on Child from $0.200$ to $0.296$, while Alzheimer's stays at or above
$0.743$.
The branch factor does not control the fairness pathway: a direct
Sex$\to$MOCA edge appears at $k{=}1$ and $k{=}5$ but not at $k{=}3$.
Hybrid GoT-CD, which adds the BIC term and correlation-weighted projection,
remains DAG-valid on all three datasets; relative to the seed-$0$ run it raises
Child from $0.222$ to $0.326$ but lowers Asia from $0.750$ to $0.571$ and
Alzheimer's from $0.757$ to $0.562$, so the data term helps only where the
prior is weak.

\textit{Query cost.}\ Pairwise querying needs $n(n{-}1)/2$ requests, $55$ on
Alzheimer's and $190$ on Child, and LLM-BFS at most $n{+}1$; GoT-CD issued
three requests in every run, consuming $3{,}221$--$3{,}628$ tokens on
Alzheimer's and $3{,}619$--$3{,}841$ on Child.

%%%%%%%%%%%%%%%%%%%%%%%%%%%%%%%%%%%%%%%%%%%%%%%%%%%%%%%%%%%%%%%%%%%%%%%%%%%%%%%%

\section{Related Work}
\label{sec:related}

\subsection{Classical causal discovery}
\label{subsec:related-classical}

Constraint-based methods such as PC~\cite{spirtes2000causation,kalisch2007pc}
and FCI~\cite{spirtes1995causal} test conditional independences and orient
edges up to Markov equivalence~\cite{glymour2019review,kitson2023survey};
score-based search such as GES~\cite{chickering2002optimal} optimizes a
Bayesian information criterion or a generalized
score~\cite{huang2018generalized}; and continuous optimization recasts
acyclicity as a smooth constraint~\cite{zheng2018dags,bello2022dagma}.
All rely on assumptions about linearity, Gaussianity, or faithfulness, and
toolkits such as causal-learn~\cite{zheng2024causal} make them standard
baselines.

\subsection{LLM-based and Graph-of-Thoughts causal discovery}
\label{subsec:related-llm}

Language models answer pairwise causal queries well above chance on standard
benchmarks, drawing on domain knowledge from pretraining rather than on
inference from data~\cite{kiciman2023causal}, and such judgments can be
assembled into full graphs~\cite{long2023causal}.
Breadth-first traversal reduces the quadratic number of queries to a linear
one by asking once for exogenous variables and once per node for its direct
effects, at the cost of committing to each expansion~\cite{jiralerspong2024efficient}.
Agentic pipelines combine LLM proposals with statistical
modules~\cite{khatibi2024alcm,abdulaal2024causal}; we instead isolate
reasoning topology and study data access separately through hybrid GoT-CD.
Recent critiques caution that classic Bayesian-network benchmarks may be
present in pretraining corpora and call for less exposed reference
graphs~\cite{srivastava2025realizing,brouillard_landscape_2025}, which is why
we pair Asia and Child with the Alzheimer's graph.
Of the branching prompting schemes~\cite{wei2022chain,yao2023tree},
Graph-of-Thoughts~\cite{besta2024got} adds aggregation over candidates, and
GoT-CD applies it with complete edge sets as thoughts.

\subsection{Causal fairness and path-specific effects}
\label{subsec:related-fairness}

Causal fairness asks through which mechanisms a protected attribute exerts
influence~\cite{makhlouf2024causality,plecko2024causal}, from counterfactual
invariance~\cite{kusner2017counterfactual} and route
constraints~\cite{kilbertus2017avoiding,nabi2018fair} to path-specific
criteria~\cite{chiappa2019path,wu2019pcfairness} built on effect
decompositions~\cite{pearl2001direct,zhou2023tracing}.
All take the graph as given, so an omitted pathway yields a confident zero.
Work on discovery for fairness has examined how discovery choices change
fairness conclusions~\cite{binkyte2023causal,zanna2025fairness}, mostly
through aggregate scores rather than recovery of the specific pathway.

%%%%%%%%%%%%%%%%%%%%%%%%%%%%%%%%%%%%%%%%%%%%%%%%%%%%%%%%%%%%%%%%%%%%%%%%%%%%%%%%

\section{Discussion}
\label{sec:discussion}

How a language model is asked for a causal graph shapes the graph it
returns.
Asked for whole graphs that are then compared, merged, and repaired, the same
model that struggles when queried pair by pair or node by node recovers the
best structure of the LLM methods, without seeing any data, and returns a DAG
in every five-seed run.
The way of asking also shapes how the graph fails, and our case study shows
that no discovery method, LLM-based or classical, reliably recovered the
pathway on which a fairness conclusion rests.

We set out to learn whether reasoning over complete candidate graphs improves
prior-informed structure recovery, and within the scope of this study, three
graphs, one backbone, and five seeds, the answer is a qualified yes.
Across five seeds, GoT-CD lowers SHD by about a quarter relative to LLM-BFS
and by roughly half to three quarters relative to pairwise querying on
Alzheimer's and Child.
The local methods fail in ways that their query structure predicts.
Pairwise querying never judges more than one pair at a time, so on Child it returns more
than a hundred edges and, in two runs, cycles; LLM-BFS commits to each node's
children before seeing the rest of the graph, and in one run it declared
every variable exogenous.
GoT-CD instead sees whole candidates side by side, merges only edges that
some candidate proposed, and projects by support, so incoherent graphs are
rejected before anything is committed.
The replay analysis supports this account: merging several candidates, rather
than the final projection, carries most of the gain.
Because the default method never touches the data, these gains belong to the
reasoning structure alone, and on Asia and Alzheimer's GoT-CD exceeds in
$F_1$ every classical baseline fitted to the same $100$ samples.

This result suggests a shift in the question that earlier work asked.
Studies of LLM causal reasoning established that the models carry usable
causal knowledge~\cite{kiciman2023causal,long2023causal}; our comparison
suggests that how the knowledge is elicited matters as much as whether it
exists, since one backbone yields markedly different graphs under pairwise,
breadth-first, and whole-graph prompting.
LLM-BFS was designed to cut the quadratic cost of pairwise
querying~\cite{jiralerspong2024efficient}, and our runs show the accuracy and
stability that its early commitment gives up; GoT-CD recovers them with only
three requests per graph.
Pipelines that pair LLM proposals with statistical
modules~\cite{khatibi2024alcm,abdulaal2024causal} assume that data evidence
improves the proposal; hybrid GoT-CD qualifies this, because a Gaussian data
term helped only on Child, where the prior is weakest, and lowered accuracy
on Asia and Alzheimer's with $100$ samples.
The gains also hold on the Alzheimer's graph, curated to be less exposed in
pretraining than the classic benchmarks~\cite{srivastava2025realizing},
although its named clinical variables may still be familiar to the model.

Better structure, however, is not the end goal when a graph is built for a
downstream analysis, and our second question was whether structural accuracy
is enough.
In our case study it is not.
The Sex$\to$MOCA Score pathway hinges on a single edge, Sex$\to$Brain Volume:
GoT-CD recovered the downstream edge Brain Volume$\to$MOCA Score in every run,
but Sex$\to$Brain Volume appeared in only one of fifteen \textsc{Generate}
candidates and in no final graph, a miss that costs one of nineteen edges in
recall and the entire effect in the fairness analysis.
Earlier work found that discovery choices change fairness
conclusions~\cite{binkyte2023causal,zanna2025fairness}; our results locate
the change in whether a specific pathway is recovered, which aggregate
scores cannot reveal.
Here, too, the way of asking shapes the error.
Queried pair by pair, the model fabricated direct edges that look like any
other edge and in one run produced a confident verdict of the wrong sign.
Queried for whole graphs, it abstained: every GoT-CD failure was a finding of
no effect, in four runs visible on inspection as a protected attribute with
no edges, and none reported a reversed disparity.
A systematic, visible omission is easier to guard against than a sporadic,
invisible fabrication, although it remains a failure.
The logged runs locate where it arises.
Every \textsc{Generate} candidate proposed at least one edge out of Sex, and
\textsc{Aggregate} kept one in four runs, including Sex$\to$Brain Volume
once; \textsc{Improve}, asked to remove weakly justified edges, then deleted
every edge out of Sex in three of those four runs.
Without \textsc{Improve}, the run that kept Sex$\to$Brain Volume would have
retained the true path, so the step that adds accuracy is also the step that
removes the demographic edge.

These answers are partial, and the design of the study sets their limits.
With one backbone at one temperature, three graphs of at most $20$ nodes, and
five seeds, our tests have little power: they indicate that whole-graph
reasoning helps in this setting, but not whether the gain persists for larger
graphs or for stronger and open-weight models, nor how often each failure
mode occurs.
The case study covers one pathway in one semi-synthetic, linear-Gaussian
graph, and its reference verdict is itself an estimate.
Fitted on the same $100$ rows, the reference effect is $-0.572$
$[-0.695,-0.440]$, and on the full $1000$-row release it is $-0.604$
$[-0.639,-0.566]$, so the negative reference verdict is stable; still, a flip
marks disagreement with an estimated reference on simulated data, not a
clinical conclusion about sex differences in cognition.
Whether structural accuracy predicts pathway recovery in general therefore
remains open, and it is hard to settle in practice, where no reference graph
exists to compute path recall; the default projection also breaks cycles by
candidate support rather than by an identification criterion.

The same whole-graph formulation that produced the omission also offers a
way to correct it.
Because GoT-CD scores and merges complete graphs, an \textsc{Improve} rule
that keeps edges out of protected attributes unless the model states a
reason to delete them, an explicit elicitation step for such edges, or a
scoring term that penalizes isolating them can be added without changing the
rest of the method.
Stronger and open-weight backbones, larger clinical graphs, further
baselines such as LiNGAM, and multi-pass Graph-of-Thoughts
loops~\cite{besta2024got} would show how far the structural gains extend.
As LLM-based discovery moves toward whole-graph and agentic pipelines, the
question worth asking of a discovered graph is not only how many edges it
gets right but whether it keeps the pathways its intended analysis depends
on.

%%%%%%%%%%%%%%%%%%%%%%%%%%%%%%%%%%%%%%%%%%%%%%%%%%%%%%%%%%%%%%%%%%%%%%%%%%%%%%%%

%%%%%%%%%%%%%%%%%%%%%%%%%%%%%%%%%%%%%%%%%%%%%%%%%%%%%%%%%%%%%%%%%%%%%%%%%%%%%%%%

% arXiv does not run BibTeX; the compiled bibliography is in main.bbl.
% To regenerate locally: replace the next line with
%   \bibliographystyle{IEEEtran}
%   \bibliography{ref}
% then run bibtex and copy the resulting .bbl back.
% Generated by IEEEtran.bst, version: 1.14 (2015/08/26)

\end{document}